\documentclass{ieeeaccess}
\usepackage{cite}
\usepackage{subfigure}
\usepackage{amsmath,amssymb,amsfonts}
\usepackage{algorithmic}
\usepackage{graphicx}
\usepackage{textcomp}
\usepackage{booktabs}
\usepackage{threeparttable}
\usepackage{multicol}
\usepackage{multirow}
\usepackage{fancyhdr}

\usepackage{indentfirst}
\def\BibTeX{{\rm B\kern-.05em{\sc i\kern-.025em b}\kern-.08em
    T\kern-.1667em\lower.7ex\hbox{E}\kern-.125emX}}
\begin{document}

\history{}
\doi{}

\title{Generation and Frame Characteristics of Predefined Evenly-Distributed Class Centroids for Pattern Classification}
\author{\uppercase{HAIPING HU}\authorrefmark{1},
\uppercase{YINGYING YAN\authorrefmark{1}, QIUYU ZHU\authorrefmark{2} and GUOHUI ZHENG}.\authorrefmark{2},
}
\address[1]{College of Sciences, ShangHai University, ShangHai 201900, China. }
\address[2]{School of Communication and Information Engineering, ShangHai University, ShangHai 201900,China }


\markboth
{Author \headeretal: Preparation of Papers for IEEE TRANSACTIONS and JOURNALS}
{Author \headeretal: Preparation of Papers for IEEE TRANSACTIONS and JOURNALS}

\corresp{Corresponding author: QIUYU ZHU (e-mail: zhuqiuyu@staff.shu.edu.cn).}
\begin{abstract}
Predefined evenly-distributed class centroids (PEDCC) can be widely used in models and algorithms of pattern classification, such as CNN classifiers, classification autoencoders, clustering, and semi-supervised learning, etc. Its basic idea is to predefine the class centers, which are evenly-distributed on the unit hypersphere in feature space, to maximize the inter-class distance. The previous method of generating PEDCC uses an iterative algorithm based on a charge model. The generated class centers will have some errors with the theoretically evenly-distributed points, and the generation time is long. This paper takes advantage of regular polyhedron in high-dimensional space and the evenly distributed points on the $n$ dimensional hypersphere to generate PEDCC mathematically. Then, we discussed the basic and extended characteristics of the frames formed by PEDCC, and some meaningful conclusions are obtained. Finally, the effectiveness of the new algorithm and related conclusions are proved by experiments. The mathematical analysis and experimental results of this paper can provide a theoretical tool for using PEDCC to solve the key problems in the field of pattern recognition, such as interpretable supervised/unsupervised learning, incremental learning, uncertainty analysis and so on.
\end{abstract}

\begin{keywords}
Predefined evenly-distributed class centroids, polyhedron in high-dimensional space, pattern classification, CNN classifiers, frame theory.
\end{keywords}

\titlepgskip=-15pt

\maketitle
\thispagestyle{plain}

\section{Introduction}
\label{sec:introduction}
 Pattern classification is to determine the class label of an input sample from a given class set. So, it is necessary to effectively extract the features of input samples based on a certain algorithm, and then classify the input samples by training the classifier. Pattern classification is widely used in modern society, especially in the fields of face recognition\cite{face1,face2},  object detection \cite{tar1,tar2}, object segmentation \cite{seg1,seg2  }, text classification\cite{text1,text3} and so on. It is an important foundation of artificial intelligence.

Pattern classification can be divided into supervised classification and unsupervised classification\cite{9}. Its essence is, through effective feature extraction, to make the features of samples of the same class gather together as much as possible, while samples of different classes are separated as much as possible. To achieve this goal, the pattern classifier is usually implemented by a loss function\cite{loss1,loss2}, in which PEDCC-Loss\cite{loss} creatively predefines the class center as a series of evenly-distributed points on the hypersphere, so that the inter-class distance reaches a maximum.

Due to the solidifying characteristics, PEDCC provides a unique research perspective for the solution of key problems in the field of pattern recognition, such as interpretable supervised/unsupervised learning, incremental learning, uncertainty analysis an so on. Now, PEDCC has been used in CNN classifiers\cite{loss}, classification autoencoders\cite{class}, clustering\cite{image}, semi-supervised learning\cite{semi}, etc. Although PEDCC has shown some excellent characteristics and has been well applied in some aspects, the mathematical generation method and related characteristics of PEDCC have not been well studied, which hinders its further application.

This paper studies the mathematical generation method of PEDCC based on the regular polyhedron in the high-dimensional space\cite{linlei}\cite{17}, analyzes its characteristics from the perspective of frame theory, and applies these characteristics to pattern classification. The main contributions of this paper are as follows:

(1) From the related properties of the regular polyhedron in the high-dimensional space, the mathematical generation method of PEDCC is given theoretically. Compared with the iterative method, this method has higher accuracy and less generation time.

(2) The related properties of the frame formed by PEDCC is discussed, and the quantitative relationship among the three angles formed by latent features, subspace spanned by PEDCC and PEDCC points is given.

(3) Based on the above mathematical analysis, for the PEDCC-Loss based CNN classifier network, experiment shows that the dimension of latent features has obvious influence on the recognition performance, although they are almost distributed on the subspace spanned by PEDCC with dimension class number-1 after training.

This article is mainly divided into six parts. The first part introduces the background and application of pattern classification. the second part is related work focusing on the original PEDCC generating method, PEDCC-Loss for CNN classifier, the regular simplex and its application on pattern classification. The third part gives the mathematical method to generate PEDCC. The forth part theoretically discusses the basic and extended characteristics of PEDCC from the perspective of the frame theory. The fifth part is the experimental comparison and verification of our method. The last part is conclusion and discussion.
\section{Related Work}
\subsection{ PEDCC and PEDCC-Loss}
PEDCC is originally generated based on the lowest charge energy physical model\cite{loss}. It uses the same polarity and the same amount of charge on the hypersphere surface. Without the influence of other factors, the points start to move continuously via the repulsive force between the charges, finally the points on the hypersphere reach a state where the charge energy is the lowest and the movement stops. At this time, the charges are the furthest away from each other and evenly distributed on the hypersphere. To generate $k$ evenly-distributed points, firstly, it need to randomly select $k$ initial predefined class centers from $n$(feature number) dimensional Gaussian distribution and normalize them, then use initialized speed parameters to describe the motion state of each point. The resultant force of each point is related to the distance between any two points, so the state of each point can be updated by its speed, and the speed can be updated by the resultant tangent vector. After continuous iteration and update of the points, these points are finally evenly distributed on the hypersphere, to ensure the maximum distance of different classes. We can manually set the class number $k$ and the feature dimension $n$, a series of randomly evenly-distributed points on the hypersphere can be generated.

PEDCC-Loss\cite{loss}, which is based on PEDCC, is a new loss function for deep learning based classifier. The values of the last full-connected layer of classifier networks are replaced by PEDCC weight. Here, the PEDCC-Loss is given as follow:
\begin{center}
\begin{align}
 L_{AM}=-\frac{1}{N}\sum_{i}\log\frac{e^{s\cdot(\cos\gamma_{y_{i}}-m)}}{e^{s\cdot(\cos\gamma_{y_{i}}-m)}+\sum^{c}_{j=1,j\neq y_{i}}e^{s\cdot\cos\gamma_{j}}}\,
\end{align}
\end{center}
\begin{align}
L_{MSE}=\frac{1}{2}\sum_{i=1}^{N}\|x_{i}-pedcc_{y_{i}}\|^{2}=\sum_{i=1}^{N}(1-\cos\gamma_{ y_{i}})^{2}
\end{align}
\begin{align}
L_{PEDCC-Loss}=L_{AM}+\lambda\sqrt[n]{L_{MSE}},
\end{align}
where $s=\|W_{i}\|\|x_{i}\|\cos\gamma_{y_{i}},  x_{i} $ is $i$th input sample, $y_{i}$ is its label, and $W_{i}$ is corresponding network weight, meanwhile, $m$ is angular margin, $\lambda$ is weighted coefficient and $N$ is class number. PEDCC-Loss is mainly constituted by improved cross entropy loss and mean square error of PEDCC with constrain factor $n$ to obtain best result in classification and face recognition tasks.

\subsection{The Regular Simplex and its Application on Pattern Classification}
In some special cases, PEDCC is a regular polytope\cite{17} in high-dimensional space, which has been proved that there are only three kinds of regular polytope in 5-dimensional space and above: simplex, hypercube and hypercross, and the number of vertices of regular simplex is dimension + 1. Due to the need for any vertex number, PEDCC needs to be obtained by proper transformation of regular polytope. Since the number of latent features is usually larger than the number of classes in pattern classification, regular simplex has been studied and applied.

For the generation of regular simplex, in\cite{19}, the author proposed a possible way of constructing $n+1$ regular simplex vertices in space $R^{n}$ :
\begin{align}
\boldsymbol{a}_{i}=
\begin{cases}
n^{-\frac{1}{2}}\boldsymbol{1}& if  i=1\\
c\boldsymbol{1}+d\boldsymbol{e}_{i-1}& if  2\leq i\leq n+1,
\end{cases}
\end{align}
where
\begin{align}
c=-\frac{1+\sqrt{n+1}}{n^{\frac{3}{2}}},d=\sqrt{\frac{n+1}{n}},
\end{align}

and $\boldsymbol{e}_{i}$ is standard unit vector with 1 in position $i$ and 0 in all other positions. By (4), we know these equidistant points are basic points, and any rotated version of these points retain the equidistant property. It is noted that only $n+1$ equidistant points can be generated in space $R^{n}$ by (4). So, we can parameter the different categories by the vertices of regular simplex to enforcing parsimony.

Based on the properties of regular simplex, Kenneth LANGE and Tong Tong Wu\cite{19}\cite{20} introduced a new method of supervised learning based on linear discrimination among the vertices of a regular simplex in Euclidean space: Vertex Discriminant Analysis(VDA). Each vertex represents a different category. Discrimination is phrased as a regression problem involving $\varepsilon-$insensitive residuals and a quadratic penalty on the coefficients of the linear predictors. Subsequently, the author put forward a new nonlinear VDA method based on reproducing kernels\cite{21}. Based on different situations, Kurnia et al.\cite{22} compared the performance of VDA with quadratic discriminant analysis(QDA) using simulated data. Recently, Vincenzo Dentamaro et al.\cite{18} proposed vertex feature classification(VFC) algorithm, used for multi-class classification, which maps input images into hyper dimensional feature space, named $'$simplex space$'$, by using multi-lateration techniques, and each class is associated with special vertex of polytope computed in the feature space.

The above mentioned methods belong to the category of traditional statistical pattern recognition, which only takes the vertices of regular simplex as the classification target, and has no effective nonlinear feature extraction method and flexibility in the number of features, so the recognition performance is limited. Because of the complicated nonlinear feature extraction by deep learning, and adaptation to any number of features and categories, the PEDCC-Loss based classifier has much better classification performance.
\section{Generation of Evenly-Distributed Points on Hypersphere}
This section gives the method of generating PEDCC mathematically, including the method of generating basic PEDCC points (vertices of simplex), and the method of PEDCC generation from basic PEDCC points.
\subsection{Generation of basic PEDCC points}
\noindent
\textbf{Proposition 1}. For arbitrarily generated $k$ points $\boldsymbol{a}_{i}(i=1,2,...,k)$ evenly-distributed on the unit hypersphere of $n$ dimensional Euclidean space, if $k\leq n+1$, such that
\begin{equation}
\langle \boldsymbol{a}_{i},\boldsymbol{a}_{j}\rangle\doteq-\frac{1}{k-1},i\neq j.
\end{equation}

One generating method of $k$ points and its proof are provided in appendix of work\cite{19}, whose equation is shown in (4),(5). In this paper, another analytical expression for constructing equidistant points on the unit hypersphere in the feature space is given in appendix A. Compared with (4), which can only generate $n+1$ equidistant points in $n$ dimensional Euclidean space(that is, the vertices of a regular simplex), we can generate any $k\leq n+1$ uniformly distributed points in high dimensional feature space(when $k=n+1$ , this is just the vertices of regular simplex). The basic PEDCC points of any dimension can also be obtained by adding 0 of the vertex vector generated by (4),(5). Although they are different, they can both obtain any random PEDCC points by the generation methods described in the next subsection.

The following is the points of the basic PEDCC generated by our analytical method when $n=4$, and $k=2,3,4,5.$
\begin{flalign}
k=2,&\boldsymbol{a}_{1}=(0,0,0,-1),\boldsymbol{a}_{2}=(0,0,0,1)&\notag\\
k=3,&\boldsymbol{a}_{1}=(0,0,-\frac{\sqrt{3}}{2},-\frac{1}{2}),\boldsymbol{a}_{2}=(0,0,\frac{\sqrt{3}}{2},-\frac{1}{2}),\notag\\
&\boldsymbol{a}_{3}=(0,0,0,1)&\notag\\
k=4,&\boldsymbol{a}_{1}=(0,-\frac{\sqrt{6}}{3},-\frac{\sqrt{2}}{3},-\frac{1}{3}),\boldsymbol{a}_{2}=(0,\frac{\sqrt{6}}{3},-\frac{\sqrt{2}}{3},-\frac{1}{3}),\notag\\
&\boldsymbol{a}_{3}=(0,0,-\frac{2\sqrt{2}}{3},-\frac{1}{3}),\boldsymbol{a}_{4}=(0,0,0,1)\notag\\
k=5,&\boldsymbol{a}_{1}=(-\frac{\sqrt{10}}{4},-\frac{\sqrt{30}}{12},-\frac{\sqrt{15}}{12},-\frac{1}{4}),\notag\\
&\boldsymbol{a}_{2}=(\frac{\sqrt{10}}{4},-\frac{\sqrt{30}}{12},-\frac{\sqrt{15}}{12},-\frac{1}{4}),&\notag\\
&\boldsymbol{a}_{3}=(0,-\frac{2\sqrt{2}}{3},-\frac{\sqrt{15}}{12},-\frac{1}{4}),\notag\\
&\boldsymbol{a}_{4}=(0,0,\frac{\sqrt{15}}{4},-\frac{1}{4}),\boldsymbol{a}_{5}=(0,0,0,1)&\notag
\end{flalign}
\subsection{Generation of arbitrary PEDCC points}
In PEDCC-Loss based classifier, the value of the same dimension of each PEDCC points can not be 0 at the same time. Otherwise, due to the solidifying characteristics of PEDCC, its gradient back propagation will be invalid. Therefore, we need generate random PEDCC points.

The following proves that for any orthogonal matrix $\boldsymbol{U},\{\boldsymbol{U}\boldsymbol{a}_{i},i=1,...,k\}$ still satisfies
\begin{align}
\langle\boldsymbol{Ua}_{i},\boldsymbol{Ua}_{j}\rangle=-\frac{1}{k-1},i\neq j.
\end{align}

For any $n$ linearly independent vectors in $n$ dimensional Euclidean space, after Schmidt orthogonalization and normalization, it is recorded as $\boldsymbol{U}$, which is an orthogonal matrix. For any $i\neq j. $
\begin{align}
\langle\boldsymbol{Ua}_{i},\boldsymbol{Ua}_{j}\rangle=(\boldsymbol{Ua}_{i})^{\top}\boldsymbol{Ua}_{j}=\boldsymbol{a}_{i}^{\top}\boldsymbol{a}_{j}=\langle \boldsymbol{a}_{i},\boldsymbol{a}_{j}\rangle=-\frac{1}{k-1}.
\end{align}

Due to the arbitrariness of the orthogonal matrix, the conclusion is completed
under such conditions. So, fixed evenly-distributed points in the space and arbitrary rotation can generate arbitrary evenly-distributed points.

Fig.1 shows the distribution of evenly-distributed points in 3-dimensional space when $k$=2,3 and 4.

\begin{figure*}[htbp]
\centering
\subfigure[]{
    \begin{minipage}[t]{0.3\linewidth}
        \centering
        \includegraphics[width=1.3in]{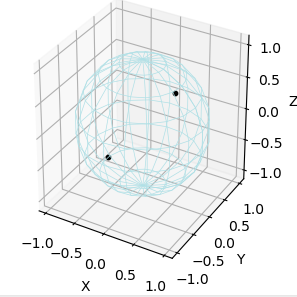}\\
        \vspace{0.02cm}

    \end{minipage}%
  }
\subfigure[]{
    \begin{minipage}[t]{0.3\linewidth}
        \centering
        \includegraphics[width=1.3in]{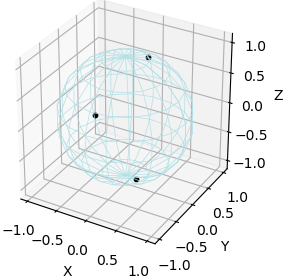}\\
        \vspace{0.02cm}

    \end{minipage}%
}
\subfigure[]{
    \begin{minipage}[t]{0.3\linewidth}
        \centering
        \includegraphics[width=1.3in]{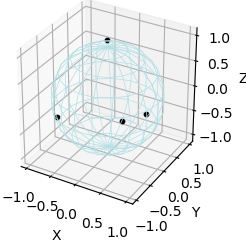}\\
        \vspace{0.02cm}

    \end{minipage}%
}
\centering
\caption{The distribution diagram of evenly-distributed points for $k$=2,3,4 in a 3-dimensional space}
\vspace{-0.2cm}
\label{image}
\end{figure*}
\section{Frame Characteristics of PEDCC}

The previous theory is mainly about the generation of evenly-distributed points. Next, we are going to discuss the application of evenly-distributed points from the frame theory.

\subsection{ Basic frame characteristics of PEDCC}
\noindent
\textbf{Proposition 2}. For the frame$\{\boldsymbol{a}_{j}\in \boldsymbol{R}^{n}|j\in J,J=1,2,...,k\}$  in the $n$ dimensional Euclidean space, the point $\boldsymbol{a}_{j}$ is evenly-distributed on the unit hypersphere and $k$ is the number of points. If $k=n+1,\forall \boldsymbol{f}\in R^{n}$, then
\begin{equation}
\sum_{1}^{k}|\langle\boldsymbol{a}_{j},\boldsymbol{f}\rangle|^{2}=(1+\frac{1}{k-1})\|\boldsymbol{f}\|^{2}.
\end{equation}
\noindent
\textbf{Proof}. when $n=1,k=2$, for any evenly-distributed unit vectors $\boldsymbol{a}_{1}, \boldsymbol{a}_{2}$ and $\boldsymbol{f}\in \boldsymbol{R}$, we have
\begin{equation}
|\langle \boldsymbol{a}_{1}, \boldsymbol{f}\rangle|^{2}+|\langle \boldsymbol{a}_{2}, \boldsymbol{f}\rangle|^{2}=2\|\boldsymbol{f}\|^{2}.
\end{equation}

Now if $k=c-1,n=c-2,\boldsymbol{a}_{j}\in \boldsymbol{R}^{c-2},j=1,2,...,c-1,$ and $\boldsymbol{f}\in \boldsymbol{R}^{c-2}$, we can get
\begin{equation}
\sum_{1}^{c-1}|\langle\boldsymbol{a}_{j}, \boldsymbol{f}\rangle|^{2}=(1+\frac{1}{c-2})\|\boldsymbol{f}\|^{2}.
\end{equation}

When $k=c,n=c-1,$  by the generation of evenly-distributed points(40), we can take any unit vector $\boldsymbol{a}_{c}$, (let $\boldsymbol{a}_{c}$ the last position is 1, and the rest are zero). Let $\boldsymbol{W}=(\boldsymbol{a}_{c})^{\top}$, we know $\boldsymbol{W}$ must be $c-2$ dimension. According to the assumption, we can get evenly-distributed points $\boldsymbol{\beta}_{1},\boldsymbol{\beta}_{2},...,\boldsymbol{\beta}_{c-1}$, which construct a tight frame and must satisfy the conclusion. meanwhile, for any $\boldsymbol{f},$
\begin{align}
\boldsymbol{f}=\sum_{1}^{c-1}\langle\boldsymbol{f},\boldsymbol{\beta}_{i}\rangle\boldsymbol{\beta}_{i},
\end{align}
$\boldsymbol{\beta}_{i}(i=1,2,...,c-1$ ) is a $c-1$ dimensional vector here, which is formed by adding zero after the $c-2$ dimension, so it is mainly considered from the low one dimension and then rises to the high one and $|\langle\boldsymbol{a}_{c},\boldsymbol{f}\rangle|$ is the absolute value of the last one dimension of $\boldsymbol{f}$. If we decompose $\boldsymbol{f}$  into the former $c-2$  dimension vector $\boldsymbol{f}_{1} $ and the last dimension $\boldsymbol{f}_{0}$, for example, $\boldsymbol{f}=(f_{1},f_{2},...,f_{c-2},f_{c-1}),\boldsymbol{f}_{1}=(f_{1},f_{2},...,f_{c-2}),{f}_{0}=(f_{c-1}) $.\\
By (11) and (12), we have
\begin{align}
  &\sum_{1}^{c-1}|\langle\boldsymbol{a}_{i},\boldsymbol{f}\rangle|^{2}=\sum_{1}^{c-1}\langle\langle\boldsymbol{a}_{i},\boldsymbol{f}\rangle,\langle\boldsymbol{a}_{i},\boldsymbol{f}\rangle\rangle\notag\\
&=\frac{c(c-2)}{(c-1)^{2}}\sum_{1}^{c-1}|\langle\boldsymbol{f},\boldsymbol{\beta}_{i}\rangle|^{2}+\frac{1}{(c-1)^{2}}\sum_{1}^{c-1}|\langle\boldsymbol{f},\boldsymbol{a}_{c}\rangle|^{2}\notag\\
  &-2\frac{\sqrt{c(c-2)}}{(c-1)^{2}}\sum_{1}^{c-1}\langle\boldsymbol{f},\boldsymbol{\beta}_{i}\rangle*\langle\sum_{1}^{c-1}\langle\boldsymbol{f},\boldsymbol{\beta}_{j}\rangle\boldsymbol{\beta}_{j},\boldsymbol{a}_{c}\rangle\notag\\
&=\frac{c}{c-1}\|\boldsymbol{f}_{1}\|^{2}+\frac{1}{c-1}|\langle\boldsymbol{a}_{c},\boldsymbol{f}\rangle|^{2},
\end{align}
\begin{align}
\sum_{1}^{c}|\langle\boldsymbol{a}_{i},\boldsymbol{f}\rangle|^{2}&=\sum_{1}^{c-1}|\langle\boldsymbol{a}_{i},\boldsymbol{f}\rangle|^{2}+|\langle\boldsymbol{a}_{c},\boldsymbol{f}\rangle|^{2}\notag\\
&=\frac{c}{c-1}\|\boldsymbol{f}_{1}\|^{2}+\frac{1}{c-1}|\langle\boldsymbol{a}_{c},\boldsymbol{f}\rangle|^{2}+|\langle\boldsymbol{a}_{c},\boldsymbol{f}\rangle|^{2}\notag\\
&=(1+\frac{1}{c-1})\|\boldsymbol{f}\|^{2}.
\end{align}

Next, we need to prove that for any orthogonal matrix $\boldsymbol{U}$, any frame formed $\{\boldsymbol{Ua}_{i},i=1,2,...k\}$ still satisfies the proposition.

\begin{align}
\sum_{1}^{k}|\langle\boldsymbol{Ua}_{i},\boldsymbol{Uf}\rangle|^{2}=(1+\frac{1}{k-1})\|\boldsymbol{f}\|^{2}.
\end{align}
\begin{align}
\sum_{1}^{k}|\langle\boldsymbol{Ua}_{i},\boldsymbol{Uf}\rangle|^{2}
&=\sum_{1}^{k}\langle\boldsymbol{Ua}_{i},\boldsymbol{Uf}\rangle*\langle\boldsymbol{Ua}_{i},\boldsymbol{Uf}\rangle\notag\\
&=\sum_{1}^{k}\boldsymbol{a}_{i}^{\top}\boldsymbol{f}*\boldsymbol{a}_{i}^{\top}\boldsymbol{f}\notag\\
&=(1+\frac{1}{k-1})\|\boldsymbol{f}\|^{2}.
\end{align}

According to the arbitrariness of $\boldsymbol{U}$, the conclusion is also established.

The proposition mainly discusses the case of $k=n+1$, that is, the number of evenly-distributed points is equal to the space dimension plus one, and the quantitative relationship is satisfied by the  projection of space vector in the frame formed by these evenly-distributed points. Next, we generalize the result.
\subsection{ Extension of PEDCC frame characteristics}
\noindent
\textbf{Proposition 3}. For the frame $\{\boldsymbol{a}_{i},j=1,2,...,k\}$ in $n$ dimensional Euclidean space, $\boldsymbol{a}_{i}$ is the evenly-distributed points on the $n$ dimensional hypersphere, and $ k$ is the number of evenly-distributed points, then for any $\forall \boldsymbol{f}\in \boldsymbol{R}^{n},c<n+1,$
\begin{align}
\sum_{1}^{k}|\langle\boldsymbol{a}_{j},\boldsymbol{f}\rangle|^{2}=(1+\frac{1}{k-1})\|\boldsymbol{f}\|^{2}\cos^{2}\alpha,
\end{align}
where $\alpha$ is the angle between $\boldsymbol{f}$ and the orthogonal projection of $\boldsymbol{f}$ on the subspace formed by the PEDCC.

\noindent
\textbf{Proof}. Since the points $\{\boldsymbol{a}_{j},j=1,2,...,k\}$  are evenly-distributed on the dimensional hypersphere, then $\boldsymbol{a}_{1}+\boldsymbol{a}_{2}+...+\boldsymbol{a}_{k}=0$, and the frame constituted by $\boldsymbol{a}_{1},\boldsymbol{a}_{2},..,\boldsymbol{a}_{k}$ is a  $k-1$ dimensional subspace. For any $\forall \boldsymbol{f}\in \boldsymbol{R}^{n}$ projecting into the $k-1$ dimensional subspace, we have $\boldsymbol{e}=\boldsymbol{f}-\boldsymbol{p}$ where $\boldsymbol{e}$ is perpendicular to the $k-1$ dimensional subspace and $\boldsymbol{p}$ ,which can be expressed linearly by the basis, is an orthogonal projection of $\boldsymbol{f}$ on the $k-1$ dimensional subspace, therefore,
\begin{align}
&\boldsymbol{p}=x_{1}\boldsymbol{a}_{1}+x_{1}\boldsymbol{a}_{2}+...+x_{k-1}\boldsymbol{a}_{k-1}=\boldsymbol{Ax},\notag\\
&\boldsymbol{A}=(\boldsymbol{a}_{1},\boldsymbol{a}_{2},...,\boldsymbol{a}_{k-1}),\boldsymbol{x}=(x_{1},x_{2},...,x_{k-1}),
\end{align}
where $x_{1},x_{2},...,x_{k-1}$ are the projection of $\boldsymbol{f}$ in the basis $\boldsymbol{a}_{1},\boldsymbol{a}_{2},...,\boldsymbol{a}_{k-1}$ and $\boldsymbol{e}$ is perpendicular to the $k-1$ dimensional space, so $\boldsymbol{f}$ is perpendicular to any vector in the $k-1$  dimensional space. We can further get $\boldsymbol{e}=\boldsymbol{f}-\boldsymbol{p}=\boldsymbol{f}-\boldsymbol{Ax}$ and
\begin{align}
 \boldsymbol{a}_{1}^{\top}(\boldsymbol{f}-\boldsymbol{Ax})=0,\boldsymbol{a}_{2}^{\top}(\boldsymbol{f}-\boldsymbol{Ax})=0,...,\boldsymbol{a}_{k-1}^{\top}(\boldsymbol{f}-\boldsymbol{Ax})=0,
\end{align}
therefore,
\begin{align}
\boldsymbol{A}^{\top}(\boldsymbol{f}-\boldsymbol{Ax})=0,
\end{align}
finally we get
\begin{align}
\boldsymbol{x}=(\boldsymbol{A}^{\top}\boldsymbol{A})^{-1}\boldsymbol{A}^{\top}\boldsymbol{f},
\end{align}
let the projection matrix is $\boldsymbol{P}$, since
\begin{align}
\boldsymbol{p}=\boldsymbol{Ax},\boldsymbol{Pf}=\boldsymbol{p},
\end{align}
we have projection matrix
\begin{align}
\boldsymbol{P}=\boldsymbol{A}(\boldsymbol{A}^{\top}\boldsymbol{A})^{-1}\boldsymbol{A}^{\top}
\end{align}
and
\begin{align}
\|\boldsymbol{p}\|^{2}=\|\boldsymbol{f}\|^{2}\cos^{2}\alpha,
\end{align}
where $\alpha$ is the angle of vector $\boldsymbol{f}$ and $\boldsymbol{p}$.

For projection $\boldsymbol{p}$, it is projected onto the frame in $k-1$ dimensional space, then
   \begin{align}
\sum_{1}^{k}|\langle\boldsymbol{a}_{i},\boldsymbol{p}\rangle|^{2}=(1+\frac{1}{k-1})\|\boldsymbol{p}\|^{2}.
\end{align}

Since $\boldsymbol{e}$ is perpendicular to any vector in the $k-1$ dimensional space, $
\boldsymbol{e}$ is perpendicular to $\{\boldsymbol{a}_{i},i=1,2,..,k\}.$ For any $\boldsymbol{f}$ projected onto the frame, we have
\begin{align}
\sum_{1}^{k}|\langle\boldsymbol{a}_{i},\boldsymbol{f}\rangle|^{2}&=\sum_{1}^{k}|\langle\boldsymbol{a}_{i},\boldsymbol{p}+\boldsymbol{e}\rangle|^{2}=\sum_{1}^{k}|\langle\langle\boldsymbol{a}_{i},\boldsymbol{p}\rangle,\langle\boldsymbol{a}_{i},\boldsymbol{p}\rangle\rangle|\notag\\
&=(1+\frac{1}{k-1})\|\boldsymbol{f}\|^{2}\cos^{2}\alpha.
\end{align}

In fact, if the angle of $\boldsymbol{a}_{i}(0\leq i\leq k)$ and $\boldsymbol{f}$ is $\gamma_{i}$, let $\boldsymbol{r}=(\cos\gamma_{1},\cos\gamma_{2},...,\cos\gamma_{k})$, we can have
 \begin{align}
\sum_{1}^{k}|\langle\boldsymbol{a}_{i},\boldsymbol{f}\rangle|^{2}&=\|\boldsymbol{f}\|^{2}\|\boldsymbol{r}\|^{2}=(1+\frac{1}{k-1})\|\boldsymbol{f}\|^{2}\cos^{2}\alpha,
\end{align}
at the same time, if $\boldsymbol{p}$ is orthogonal projection of $\boldsymbol{f}$ and the angle between the $\boldsymbol{a}_{i}(0\leq i\leq k)$ and $\boldsymbol{f}$ is $\beta_i$, let $\boldsymbol{b}=(\cos\beta_{1},\cos\beta_{2},...,\cos\beta_{k})$, we can have
 \begin{align}
\sum_{1}^{k}|\langle\boldsymbol{a}_{i},\boldsymbol{p}\rangle|^{2}=\|\boldsymbol{p}\|^{2}\|\boldsymbol{\beta}\|^{2}=(1+\frac{1}{k-1})\|\boldsymbol{p}\|^{2},
\end{align}
then we  have
\begin{align}
\|\boldsymbol{r}\|=\sqrt{1+\frac{1}{k-1}}|\cos\alpha|=\|\boldsymbol{b}\||\cos\alpha|,
\end{align}
and for any $\boldsymbol{a}_{i},i={1,2,..,k}$, $\|\boldsymbol{p}\|=\|\boldsymbol{f}\||\cos\alpha|$, according to the law of cosine, we can have\\
\begin{align}
 \|\boldsymbol{a}_{i}-\boldsymbol{p}\|=\|\boldsymbol{f}\|^{2}\cos\alpha^{2}+2\|\boldsymbol{f}\||\cos\alpha |\cos\beta_{i}+1,
\end{align}
by $\boldsymbol{e}$ and $\boldsymbol{a}_{i}-\boldsymbol{p}$, then
\begin{align}
 \|\boldsymbol{e}-(\boldsymbol{a}_{i}-\boldsymbol{p})\|=\|\boldsymbol{f}\|^{2}+2|\cos\alpha| \cos\beta_{i}+1,
\end{align}
 $\boldsymbol{f}$ and $\boldsymbol{a}_{i}$, we can have
\begin{align}
\|\boldsymbol{f}-\boldsymbol{a}_{i}\| =\|\boldsymbol{e}-(\boldsymbol{a}_{i}-\boldsymbol{p})\|=\|\boldsymbol{f}\|^{2}+2\cos\gamma_{i}+1,
\end{align}
according to the properties of the projection matrix and frame theory, so, we can get
\begin{align}
 \cos\gamma_{i}=\cos\beta_{i}\cos\alpha,i={i,2,...,k}.
\end{align}

\begin{figure}[bh]
\centerline{\includegraphics[height=3cm,width=5.5cm]{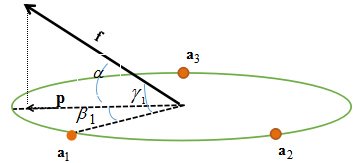}}
\vspace*{8pt}
\caption{Geometric schematic diagram of frame theory in 3 dimensional space}
\label{image2}
\end{figure}

The following is the geometric meaning of this theory in three-dimensional space, as shown in Fig.\ref{image2}. Here $\boldsymbol{a}_{1},\boldsymbol{a}_{2},\boldsymbol{a}_{3}$ constitute the frame and $\boldsymbol{p}$ is orthogonal projection of  $\boldsymbol{f}$.

After the generalization of this proposition, where we mainly discussed the properties of orthogonal projection and projection matrix under $k<n+1$ , the quantitative relationship satisfied by any vector and its frame can be expressed through orthogonal projection.

From the above theoretical analysis, we know that $c$ PEDCC points constitute a tight frame which can span a $c-1$ dimensional subspace, and any $c-1$ PEDCC points are the Riesz basis of the subspace.

\section{ experiment and verification}

To study the practical significance of the above theories, we conduct experiments to verify them respectively. Here, our experiment, which is implemented using Pytorch on an Inter(R)i7-6700CPU, 32GB RAM, and a Nvidia GTX 1080 Ti GPU, performs new algorithm and iterative method on generating speed, recognition rate and Euclidean distance. Meanwhile, by means of PEDCC-Loss based CNN classifier, we verified the application of the above theoretical analysis.

\subsection{ Comparison of iterative PEDCC algorithm and new algorithm}

After the theoretical proof of evenly-distributed points, we hope to compare the iterative method with the new algorithm. We conducted three experimental verifications to obtain the performance of iterative method and new algorithm in different context.\\
(1) Comparison of PEDCC generation speed

In a 300-dimensional space, the number of points is 50, 100, 150, 200 in Table 1; when the number of points is 100, the space dimension is 200, 300, 400, 500 in Table 2.

\begin{table}
\centering
\caption{The time (second) that two algorithms take to generate 50,100,150,200 evenly-distributed points in 300 dimensional spaces}
\label{table}
\setlength{\tabcolsep}{3pt}
\begin{tabular}{@{}c|cccc@{}}
\hline
\hline
Point number &50 & 100 & 150&200 \\\hline
Iterative method &1410.40s& 1384.04s& 1443.60s& 1279.62s \\\hline
New algorithm &2.086s & 2.80s & 3.62s & 4.52s\\
\hline
\hline
\end{tabular}
\label{1}
\end{table}

\begin{table}
\centering
\caption{The time (second) that two algorithms take to generate 100 evenly-distributed points in 200, 300, 400 and 500 dimensional spaces}
\label{table}
\setlength{\tabcolsep}{3pt}
\begin{tabular}{@{}c|cccc@{}}

\hline
\hline
Dimension &200 & 300 & 400&500 \\ \hline
Iterative method &126.50s& 137.97s& 160.00s& 154.90 s\\\hline
New algorithm &1.31s &2.67s &4.81s & 9.57s\\
 \hline
 \hline
\end{tabular}
\label{2}
\end{table}
\noindent
(2) The cosine distance between PEDCC points

Cosine distance is used as a measurement tool to compare the points generated by the new algorithm with iterative one. Here we let the number of classes $k=10$ and the feature dimension $n=1000$. For the evenly-distributed points generated by iterative PEDCC method and the new algorithm, we can calculate the distance of the points (because the distance distribution matrix is symmetric, only the half of the table is displayed for the convenient observation). Theoretically, the sum of all PEDCC points should be equal to zero, that is, the hyperplane spanned by PEDCC passes through the origin. However, the sum of the points generated by the iterative algorithm may not be equal to zero. Then, the origin is not on the the hyperplane formed by the PEDCC, so the angle between the points will be less than the theoretical value. Comparing Table 3 with Table 4, we can find that the cosine distance between any two points generated by the new algorithm is equal, which overcomes the error caused by iterative method.

It should be noted here that the conclusion is established under the premise of $k\leq n+1$, however, when $k>n+1$, this property is not established.
\begin{table}
\centering
\caption{Cosine distance table between any two points generated by the new algorithm}
\label{table}
\setlength{\tabcolsep}{3pt}
\begin{tabular}{@{}c|cccccccccc@{}}
\hline
\hline
    & 0 & 1 & 2 & 3 & 4 & 5 & 6 & 7 & 8 & 9  \\ \hline
0 &0&-0.11&-0.11&-0.11&-0.11&-0.11&-0.11&-0.11&-0.11&-0.11\\
1 &0&0&-0.11&-0.11&-0.11&-0.11&-0.11&-0.11&-0.11&-0.11\\
2 &0&0&0&-0.11&-0.11&-0.11&-0.11&-0.11&-0.11&-0.11\\
3 &0&0&0&0&-0.11&-0.11&-0.11&-0.11&-0.11&-0.11\\
4 &0&0&0&0&0&-0.11&-0.11&-0.11&-0.11&-0.11\\
5 &0&0&0&0&0&0&-0.11&-0.11&-0.11&-0.11\\
6 &0&0&0&0&0&0&0&-0.11&-0.11&-0.11\\
7 &0&0&0&0&0&0&0&0&-0.11&-0.11\\
8 &0&0&0&0&0&0&0&0&0&-0.11\\
9 &0&0&0&0&0&0&0&0&0&0\\
\hline
\hline
\end{tabular}
\label{4}
\end{table}
\begin{table}
\centering
\caption{Cosine distance table between any two points generated by the iterative PEDCC algorithm}
\label{table}
\setlength{\tabcolsep}{2pt}
\begin{tabular}{@{}c|cccccccccc@{}}
\hline
\hline
 & 0 & 1 & 2 & 3 & 4 & 5 & 6 & 7 & 8 & 9  \\ \hline
0 &0&-0.035&-0.084&-0.075&-0.012&-0.051&0.056&0.032&0.004&-0.020\\
1 &0&0&0.037&-0.003&-0.009&-0.027&-0.029&-0.015&0.023&0.006\\
2 &0&0&0&0.019&-0.002&0.016&0.002&-0.041&0.016&-0.022\\
3 &0&0&0&0&0.019&0.015&-0.004&-0.045&-0.050&0.005\\
4 &0&0&0&0&0&-0.055&-0.047&-0.042&-0.046&0.002\\
5 &0&0&0&0&0&0&-0.076&-0.045&0.005&-0.003\\
6 &0&0&0&0&0&0&0&0.040&-0.028&0.011\\
7 &0&0&0&0&0&0&0&0&-0.010&0.008\\
8 &0&0&0&0&0&0&0&0&0&-0.005\\
9 &0&0&0&0&0&0&0&0&0&0\\
\hline
\hline
\end{tabular}
\label{5}
\end{table}
\subsection{Experimental comparison of PEDCC frame characteristics}
The experiment mainly verifies whether the hidden features of both the training set and the test set fall on the frame formed by PEDCC, with Cifar10 and Cifar100 dataset on the CNN classifier. The role of PEDCC-Loss is to make $\cos\gamma_{i}$ close to 1, to minimize the misclassification. So we need $\cos\beta_{i}$ and $\cos\alpha$ close to 1 by (33). The experimental result in Table 5 shows that the angle of between the latent features and subspace spanned by PEDCC is nearly zero, which means that although the latent features have large dimension, they are almost distributed on the subspace spanned by PEDCC with dimension class number-1 after training.
\begin{table}
\centering
\caption{The average angle(degree) between the hidden features and PEDCC frame for the training set and the test set in Cifar10 and Cifar100 datasets}
\label{table}
\setlength{\tabcolsep}{3pt}
\begin{tabular}{@{}c|cc@{}}

\hline
\hline
  &Training set & Test set \\ \hline
Cifar 10 &0.09841& 0.1305 \\
Cifar 100 &0.03761 &0.0631 \\
 \hline
 \hline
\end{tabular}
\label{tab1}
\end{table}

\subsection{Performance comparison of CNN classifier based on PEDCC-Loss}
\noindent(1) Performance in PEDCC-Loss classifier

Meanwhile, we use the PEDCC-Loss function of the CNN classifier as the research object with Resnet-50, to verify the performance of the new method. Here Cifar10, Cifar100, Tiny Imagenet data sets are used in the experiment.

(i) The Cifar10 dataset has 60,000 images and total of 10 classes, and each image has a pixel size of 32*32. Every class has 5000 training images and 1000 test images.

(ii) The Cifar100 data set has 70,000 images and 100 categories in total. The size of each picture is a 32$\ast$32, and each class has 600 images, in which 500 images are training set and 100 images are test set.

(iii) The Tiny Imagenet dataset has 100,000 training images and 10,000 test images, and each image is 64$\ast$64. There are a total of 200 classes, and each class has 500 training images, 50 verification images and 50 test images.

Here we respectively set the class number of 10 and the feature number of 256 in Cifar10, 100 classes and feature number 512 in Cifar100, and 10 classes and feature number 512 in Tiny Imagenet to verify experiment.

The average recognition rate in three experiment in Table 6 show that new method of mathematic generation has slight improvement in recognition performance. 
\begin{table}
\centering
\caption{The average recognition rate between Iterative method and New method  in Cifar10, Cifar100 and Tiny Imagenet datasets}
\label{table}
\setlength{\tabcolsep}{3pt}
\begin{tabular}{@{}c|ccc@{}}

\hline
\hline
  &Cifar10 & Cifar100&Tiny Imagenet\\ \hline
Iterative method &93.83\%& 73.07\%&59.82\% \\
New method &93.92\% &73.01\%& 59.85\%\\
 \hline
 \hline
\end{tabular}
\label{tab1}
\end{table}

\noindent(2) The influence of different dimensions of latent features on the recognition rate

Although we know the features are distributed in the subspace spanned by PEDCC after training in subsection $B$, do we need only $c-1$ dimensional latent features to classify well? We take Cifar 10 data set as an example to compare the influence of latent feature number as shown in Table 7. It is found that the number of latent features has an obvious influence on the recognition results, and there is an optimal dimension. This is because, in the training process, the hidden features of the samples are not distributed in the PEDCC subspace, which makes the network be able to extract more effective nonlinear features. However, when the dimension is greater than 256, the recognition rate is basically unchanged, which indicates that more dimension are not always conducive to the improvement of recognition rate.

The influence of the number of latent features on recognition results is more obvious when the class is less, which also shows the advantage of PEDCC method which can generate any number of class center points.

\begin{table}
\centering
\caption{ The impact of dimensions of latent features on the recognition rate with CIFAR10 data set}
\label{table}
\setlength{\tabcolsep}{1.5pt}
\begin{tabular}{@{}c|cccccccc@{}}

\hline
\hline
Dimension&9 &10&20&30&50&256&512&1024\\ \hline
Reco.Rate(\%) &93.05& 92.76&93.51 &93.77&93.26&93.85&93.71&93.97\\

 \hline
 \hline
\end{tabular}
\label{tab1}
\end{table}

\section{Conclusion and Discussion}

 This paper mainly studies the construction of evenly-distributed points in high-dimensional space and the corresponding theoretical position relationship to avoid errors caused by iteratively generating evenly-distributed points of PEDCC. At the same time, from the perspective of the frame theory, we take the PEDCC as a frame, and discuss its basic and extended frame characteristics, including the quantitative relationship satisfied by the projection of the vector in the feature space onto the frame formed by PEDCC. Finally, experiment shows that the new algorithm is much faster than the iterative method, and the positions are completely accurate. Meanwhile, although the latent features of the samples of the CNN classifier are basically distributed in a subspace formed by PEDCC with the dimension of class number-1, the latent features with higher dimension are still favorable for recognition performance.

The theory discussed in this article is applicable to $k\leq n+1$ , that is, the number of points is less than or equal to the feature dimension +1. Although this situation caters to the actual situation of most classifiers. For $k>n+1$ it still is a further research topic. In the future, based on PEDCC and its frame characteristics, we will first study the uncertainty of classifiers, and further study the interpretable supervised/unsupervised learning, incremental learning and other key problems of pattern recognition.
\section{appendix A}
\noindent
\textbf{Proposition 1}. For arbitrarily generated $k$ points $\boldsymbol{a}_{i},i=1,2,...,k$ evenly-distributed on the unit hypersphere of $n$ dimensional Euclidean space, if $k\leq n+1$, such that
\begin{equation}
\langle \boldsymbol{a}_{i},\boldsymbol{a}_{j}\rangle\doteq-\frac{1}{k-1},i\neq j,
\end{equation}
\textbf{Proof}. Since $k$ points are evenly-distributed, there must be any $k-1$ points to form a $k-1$ dimensional subspaces, satisfying $\boldsymbol{a}_{1}+\boldsymbol{a}_{2}+...+\boldsymbol{a}_{k}=0$. We can assume that $\boldsymbol{a}_{1},\boldsymbol{a}_{2},...,\boldsymbol{a}_{k-1}$ is the basis. Otherwise, if it is linearly related, there must exist a set of real numbers  that are not all zero $m_{1},m_{2},...,m_{k-1}$, let $m_{j},m_{z}\neq 0,1\leq j,z \leq k-1$, satisfying
\begin{equation}
m_{1}\boldsymbol{a}_{1}+m_{2}\boldsymbol {a}_{2}+...+m_{k-1}\boldsymbol{a}_{k-1}=0,
\end{equation}
we have
\begin{equation}
\boldsymbol{a}_{j}=-\frac{m_{1}}{m_{j}}\boldsymbol{a}_{1}-...-\frac{m_{j-1}}{m_{j}}\boldsymbol{a}_{j-1}-\frac{m_{j+1}}{m_{j}}\boldsymbol{a}_{j+1}-...-\frac{m_{k-1}}{m_{j}}\boldsymbol{a}_{k-1},
\end{equation}
because of evenly distributed points and $(36)$, for $\forall m_{i}\neq m_z\neq m_{j}$

\begin{equation}
\begin{aligned}
 \langle \boldsymbol{a}_{i},\boldsymbol{a}_{j}\rangle = -\frac{m_{i}}{m_{j}}-\left( {\frac{m_{1}+...m_{i-1}+m_{i+1}+...}{m_{j}}} \right. \\
 \left. {+\frac{+m_{j-1}+m_{j+1}+...m_{k-1}}{m_{j}}} \right)\langle \boldsymbol{a}_{i},\boldsymbol{a}_{j}\rangle,
 \end{aligned}
\end{equation}
and
\begin{equation}
\begin{aligned}
 \langle \boldsymbol{a}_{z},\boldsymbol{a}_{j}\rangle = -\frac{m_{z}}{m_{j}}-\left( {\frac{m_{1}+...m_{z-1}+m_{z+1}+...}{m_{j}}} \right. \\
 \left. { +\frac{m_{j-1}+m_{j+1}+...m_{k-1}}{m_{j}}} \right)\langle \boldsymbol{a}_{z},\boldsymbol{a}_{j}\rangle,
 \end{aligned}
\end{equation}

by$(37)-(38)$, we have
\begin{equation}
(m_{z}-m_{i})\langle \boldsymbol{a}_{i}, \boldsymbol{a}_{z}\rangle=m_{z}-m_{i},
\end{equation}
therefore, $\langle \boldsymbol{a}_{i},\boldsymbol{a}_{z}\rangle=1$, it is obviously contradictory, meanwhile, $\boldsymbol{a}_{k}$ can be expressed linearly by $\boldsymbol {a}_{1},\boldsymbol {a}_{2},...,\boldsymbol {a}_{k-1},$ so $\boldsymbol {a}_{1},\boldsymbol {a}_{2},...,\boldsymbol {a}_{k-1}$ is a set of basics of $k-1$ dimensional space. Since each subspace of finite-dimensional Euclidean space has orthogonal complementary space, there is $n-k+1$ dimensional orthogonal complementary space.

We perform mathematical induction on it, and here $k\geq 2,k,n\in N$ is obvious.

When $  n=2,k=2,$ we can set $\boldsymbol {a}_{1}$ as an any unit vector, there is a vector $\boldsymbol {a}_{2}$ satisfying$\langle\boldsymbol {a}_{1},\boldsymbol {a}_{2}\rangle=-1 $. If $n=2,k=3,\boldsymbol {a}_{1}$ is an arbitrary unit vector in the space V, then there must exist vectors $\boldsymbol {a}_{2},\boldsymbol {a}_{3}$, satisfying$\langle\boldsymbol {a}_{i},\boldsymbol {a}_{j}\rangle=-\frac{1}{2},i\neq j.$

If $n=m-1$, for any $k(k\leq m)$ evenly-distributed points on the hypersphere, $\langle\boldsymbol {a}_{i},\boldsymbol {a}_{j}\rangle=-\frac{1}{k-1},i\neq j$ is satisfied.

 When $n=m,$  we can take an arbitrary unit vector $ \boldsymbol {a}_{k}(k\leq m+1)$  and consider $W=(\boldsymbol {a}_{k})^{\bot}$as an $m-1 $ dimensional Euclidean space. By assumption, the $k-1$ evenly-distributed points satisfy $\langle\boldsymbol {a}_{i}, \boldsymbol {a}_{j}\rangle=-\frac{1}{k-2}, i\neq j, k-1\leq m$, and we can have $k-1$ evenly-distributed vectors $\boldsymbol{\beta}_{1}, \boldsymbol{\beta}_{2},...,\boldsymbol{\beta}_{k-1}, \boldsymbol{\beta}_{i}$  is the orthogonal projections of $\boldsymbol{a}_{i}$. For each $i=1,2,...,k-1$, we can have
\begin{equation}
\boldsymbol{a}_{i}=\frac{\sqrt{k(k-2)}}{k-1}\boldsymbol{\beta}_{i}-\frac{1}{k-1}\boldsymbol{a}_{k},i=1,2,..,k-1,
\end{equation}
by $(40)$, we have
\begin{align}
\langle\boldsymbol{a}_{i}, \boldsymbol{a}_{j}\rangle 
&=\frac{k(k-2)}{(k-1)^{2}}\langle\boldsymbol{\beta}_{i}, \boldsymbol{\beta}_{j}\rangle+\frac{1}{(k-1)^{2}}\langle\boldsymbol{a}_{k}, \boldsymbol{a}_{k}\rangle
\notag\\&=-\frac{1}{k-1}(i\neq j,1\leq i,j\leq k-1),
\end{align}
\begin{align}
\langle \boldsymbol{a}_{i}, \boldsymbol{a}_{k}\rangle=\langle\frac{\sqrt{k(k-2)}}{k-1}\boldsymbol{\beta}_{i}-\frac{1}{k-1}\boldsymbol{a}_{k}, \boldsymbol{a}_{k}\rangle=-\frac{1}{k-1},
\end{align}
consequently, we can obtain a fixed $\boldsymbol{a}_{1},\boldsymbol{a}_{2},..,\boldsymbol{a}_{k}.$

\begin{IEEEbiography}[{\includegraphics[width=1in,height=1.25in,clip,keepaspectratio]{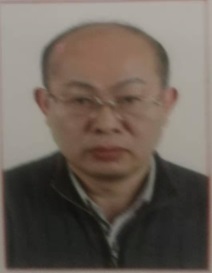}}]{haiping hu} received his Bachelor and Master degree
from Anhui Normal University in 1989, and Harbin Institute of Techniology 1992
respectively. In 2002, he received his PHD Degree in information and communication
engineering from Shanghai University. Now, he is the associate professor in Shanghai
University. His research interests include image processing, computer vision, machine
learning,Optimization etc. He is a coauthor of approximately 30 academic papers.
\end{IEEEbiography}

\begin{IEEEbiography}[{\includegraphics[width=1in,height=1.25in,clip,keepaspectratio]{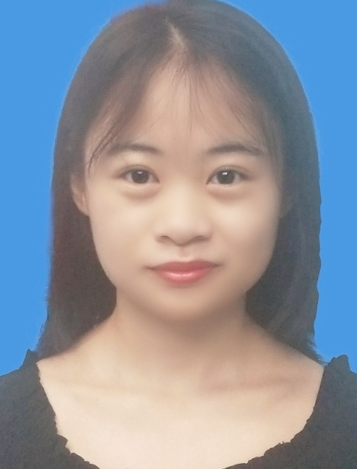}}]{yingying yan}  received the bachelor's degree from XinYang
Normal University in 2019,she is currently  pursuing the master's degree with the
College of sciences, ShangHai University. Her research  major in computational mathematics.
\end{IEEEbiography}

\begin{IEEEbiography}[{\includegraphics[width=1in,height=1.25in,clip,keepaspectratio]{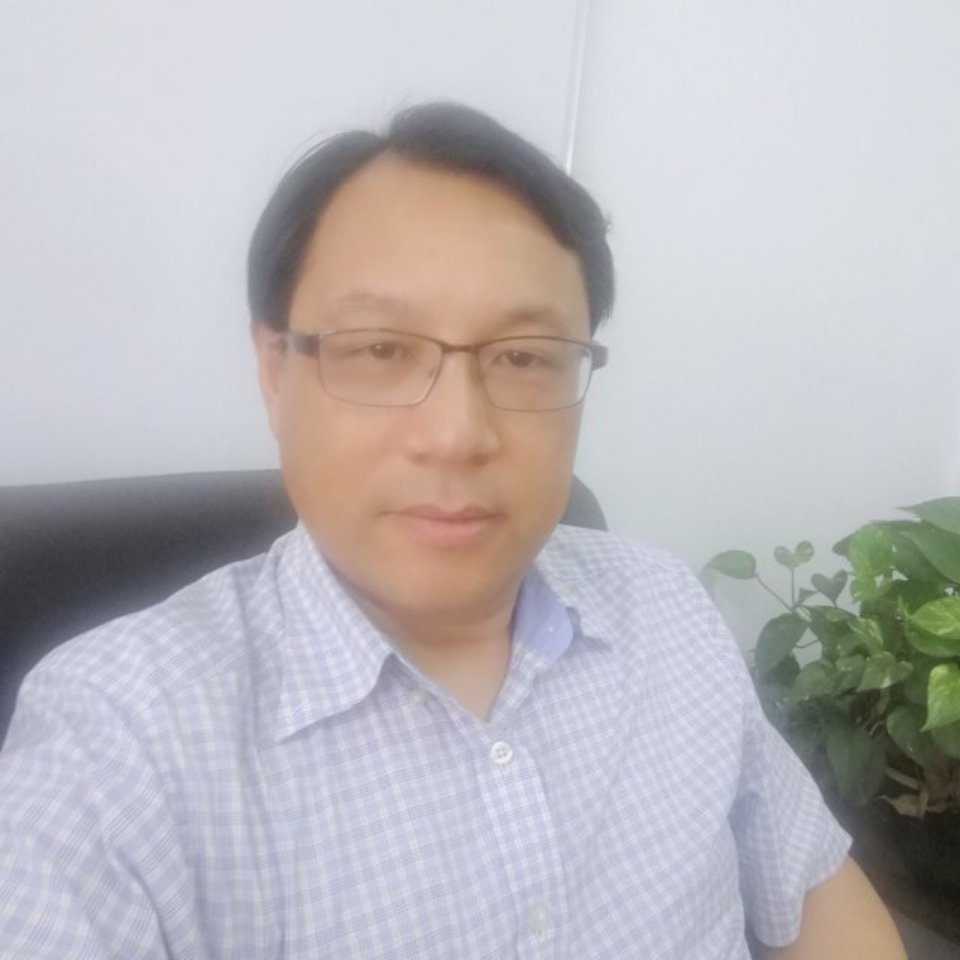}}]{qiuyu zhu} received his Bachelor and Master degree from FUDAN University
in 1985,and Shanghai University of Science and Technology in 1988
respectively. In 2006, he received his PHD Degree in information
and communication engineering from Shanghai University. Now, he is
the professor in Shanghai University.His research interests include
image processing, computer vision, machine learning, smart city, computer
application, etc. He is a coauthor of approximately 100 academic papers, and
principal investigator for more than 10 governmental funded research projects,
more than 30 industrial research projects, many of which have been widely applied.
\end{IEEEbiography}

\begin{IEEEbiography}[{\includegraphics[width=1in,height=1.25in,clip,keepaspectratio]{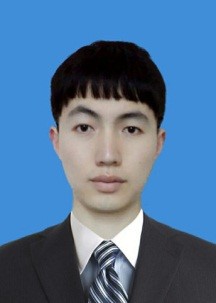}}]{guohui zheng}is currently  pursuing School of Communication and
Information Engineering, Shanghai University, Shanghai, China.He is
a master student with a research field of computer vision.
\end{IEEEbiography}

\EOD

\end{document}